\def\year{2020}\relax
\documentclass[letterpaper]{article} 
\usepackage{aaai20}  
\usepackage{times}  
\usepackage{helvet} 
\usepackage{courier}  
\usepackage[hyphens]{url}  
\usepackage{graphicx} 
\usepackage{graphicx}  
\usepackage{microtype}      
\usepackage{algorithm}
\usepackage{algorithmic}
\usepackage{amsmath}
\usepackage{amssymb}
\DeclareMathOperator*{\argmax}{argmax} 
\title{Reliability Does Matter: An End-to-End Weakly Supervised Semantic Segmentation Approach}
\author{Bingfeng Zhang,\textsuperscript{\rm 1}
	Jimin Xiao,\textsuperscript{\rm 1}\thanks{Corresponding author}
	Yunchao Wei,\textsuperscript{\rm 2}
	Mingjie Sun,\textsuperscript{\rm 1} 
	Kaizhu Huang\textsuperscript{\rm 1 3} \\
	\textsuperscript{\rm 1}Xi'an Jiaotong-liverpool University, Suzhou, China\\ \textsuperscript{\rm 2}University of Technology Sydney, Australia\\
	\textsuperscript{\rm 3}Alibaba-Zhejiang University Joint Institute of Frontier Technologies, Hangzhou, China\\
	\textsuperscript{\rm 1}\{Bingfeng.Zhang, jimin.xiao, mingjie.sun18, kaizhu huang\}@xjtlu.edu.cn, 
	\textsuperscript{\rm 2}yunchao.wei@uts.edu.au 
}

\begin{document}

\maketitle

\begin{abstract}
Weakly supervised semantic segmentation is a challenging task as it only takes image-level information as supervision for training but produces pixel-level predictions for testing. To address such a challenging task, most recent state-of-the-art approaches propose to adopt two-step solutions, \emph{i.e. } 1) learn to generate pseudo pixel-level masks, and 2) engage FCNs to train the semantic segmentation networks with the pseudo masks. However, the two-step solutions usually employ many bells and whistles in producing high-quality pseudo masks, making this kind of methods complicated and inelegant. In this work, we harness the image-level labels to produce reliable pixel-level annotations and design a fully end-to-end network to learn to predict segmentation maps. Concretely, we firstly leverage an image classification branch to generate class activation maps for the annotated categories, which are further pruned into confident yet tiny object/background regions. Such reliable regions are then directly served as ground-truth labels for the parallel segmentation branch, where a newly designed dense energy loss function is adopted for optimization. Despite its apparent simplicity, our one-step solution achieves competitive mIoU scores (\emph{val}: 62.6, \emph{test}: 62.9) on Pascal VOC compared with those two-step state-of-the-arts. By extending our one-step method to two-step, we get a new state-of-the-art performance on the Pascal VOC
(\emph{val}: 66.3, \emph{test}: 66.5).

\end{abstract}

\section{Introduction}
Recently, weakly supervised semantic segmentation receives great interest and is being extensively studied. Requiring merely low degree (cheaper or simpler) annotations including scribbles \cite{lin2016scribblesup,vernaza2017learning,tang2018regularized}, bounding boxes \cite{dai2015boxsup,khoreva2017simple}, points \cite{maninis2018deep,bearman2016s} and image-level labels \cite{ahn2018learning,hou2018self,wei2018revisiting} for training, weakly supervised  semantic segmentation offers a much easy way than its fully supervised counterpart adopting pixel-level masks~\cite{chen2018deeplab,chen2017rethinking,long2015fully}. Among these weakly supervised labels, the image-level annotation is the simplest one to collect yet also the most challenging case since there is no direct mapping between semantic labels and pixels. 

To learn semantic segmentation models using image-level labels as supervision, many existing approaches can be categorized as one-step approaches and two-step approaches. One-step approaches~\cite{papandreou1502weakly} often establish an end-to-end framework, which augments multi-instance learning with other constrained strategies for optimization. This family of methods is elegant and easy to implement. However, one significant drawback of these approaches is that the segmentation accuracy is far behind their fully supervised counterparts. To achieve better segmentation performance, many researchers alternatively propose to leverage two-step approaches~\cite{wei2017object,huang2018weakly}. This family of approaches usually aim to take bottom-up~\cite{hou2017deeply} or top-down~\cite{zhang2018top,zhou2016learning} strategies to firstly generate high-quality pseudo pixel-level masks with image-level labels as supervision. These pseudo masks then act as ground-truth and are fed into the off-the-shelf fully convolutional networks such as FCN~\cite{long2015fully} and Deeplab~\cite{chen2014semantic,chen2018deeplab} to train the semantic segmentation models. Current state-of-the-arts are mainly two-step approaches, with segmentation performance approaching that of their fully supervised counterparts. However, to produce high-quality pseudo masks, these approaches often employ many bells and whistles, such as introducing additional object/background cues from object proposals~\cite{pinheiro2015image} or saliency maps \cite{jiang2013salient} in an off-line manner. Therefore, the two-step approaches are usually very complicated and hard to be re-implemented, limiting their application to research areas such as object localization and video object tracking. 

In this paper, we present a simple yet effective one-step approach, which can be easily trained in an end-to-end manner. It achieves competitive segmentation performance compared with two-step approaches. Our approach named \emph{Reliable Region Mining} (\emph{RRM}) includes two branches: one to produce pseudo pixel-level masks using image-level annotations, and the other to produce the semantic segmentation results. In contrast to the previous two-step state-of-the-arts~\cite{ahn2018learning,lee2019ficklenet} that prefer to mine dense and integral object regions, our \emph{RRM} only chooses those confident object/background regions that are usually tiny but with high response scores on the class activation maps. We find these regions can be further pruned into more reliable ones by augmenting an additional CRF operation, which are then employed as supervision for the parallel semantic segmentation branch. With limited pixels as supervision, we designed a regularized loss named dense energy loss, which cooperates with the pixel-wise cross-entropy loss to optimize the training process. 

Despite its apparent simplicity, our one-step \emph{RRM} achieves 62.6 and 62.9 of mIoU scores on the Pascal VOC \emph{val} and \emph{test} sets, respectively. These results achieve state-of-the-art performance and it is even competitive compared with those two-step state-of-the-arts, which usually adopt complex bells and whistles to produce pseudo masks. We believe that our proposed \emph{RRM} offers a new insight to the one-step solution for weakly supervised semantic segmentation. Besides, in order to show the effectiveness of our method, we also extend our method to a two-step framework and  get a new state-of-the-art performance with 66.3 and 66.5 on the Pascal VOC \emph{val} and \emph{test} sets. Code will be made publicly available.

\section{Related Work}

Semantic segmentation is an important task in computer vision \cite{wei2018revisiting,xiao2019ian,xie2018correlation}, which requires to predict pixel-level classification.  Long \emph{et al.}~\cite{long2015fully} proposed the first fully convolutional network for semantic segmentation. Chen \emph{et al.}~\cite{chen2014semantic} proposed a new deep neural network structure named "Deeplab" to conduct pixel-wise prediction using atrous convolution, and a series of new network structures was developed after that   \cite{chen2018deeplab,chen2017rethinking,chen2018encoder}. However, fully supervised semantic segmentation requires dense pixel-level annotations, which cost expensive human expense. Weakly supervised semantic segmentation has been drawing much attention as less human intervention is needed. There are different categories of weakly supervised semantic segmentation based on the types of supervision: scribble \cite{tang2018normalized,lin2016scribblesup}, bounding box \cite{song2019box,hu2018learning,rajchl2017deepcut}, point \cite{maninis2018deep,bearman2016s} and image-level class label \cite{zhang2018adversarial,vernaza2017learning,zhang2018self}.
In this paper, we focus on image-level supervised semantic segmentation.

Image-level weakly supervised semantic segmentation only provides image-level annotation. Most recent approaches are based on class activation map (CAM) \cite{zhou2016learning}, which is to generate initial object seeds or regions from image-level annotation. Such initial object seeds or regions are converted to generate pseudo labels to train a semantic segmentation model. Wei \emph{et al.}~\cite{wei2017object} proposed to  erase iteratively the discriminative areas computed by a classification network so that more seed regions can be mined which are then combined with a saliency map to generate the pseudo pixel-level label. Wei \emph{et al.}~\cite{wei2018revisiting} also proved that dilated convolution can increase the receptive filed and improve the weakly segmentation network performance. Besides, Wang \emph{et al.}~\cite{wang2018weakly} trained a region network and a pixel network to make prediction from image level to region level and from region level to pixel level gradually. Also, this method takes saliency map as extra supervision. Ahn and Suha~\cite{ahn2018learning}  designed an affinity network to compute the relationship between different image pixels and exploited this network to get the pseudo object labels for segmentation model training. Huang \emph{et al.}~\cite{huang2018weakly} deployed a traditional algorithm named seed growing to iteratively expand the seed regions. 

However, all the above methods produced high-quality pseudo masks using a wide varieties of techniques, meaning that we need at least one or two extra networks before training FCNs for semantic segmentation prediction. In this work, we try to design one single network for the whole task to simplify the process. We believe this work offers a new perspective for the image-level weakly supervised semantic segmentation task.

\section{Proposed Method}
\begin{figure*}\
	\centering
	\includegraphics[width=0.95\textwidth]{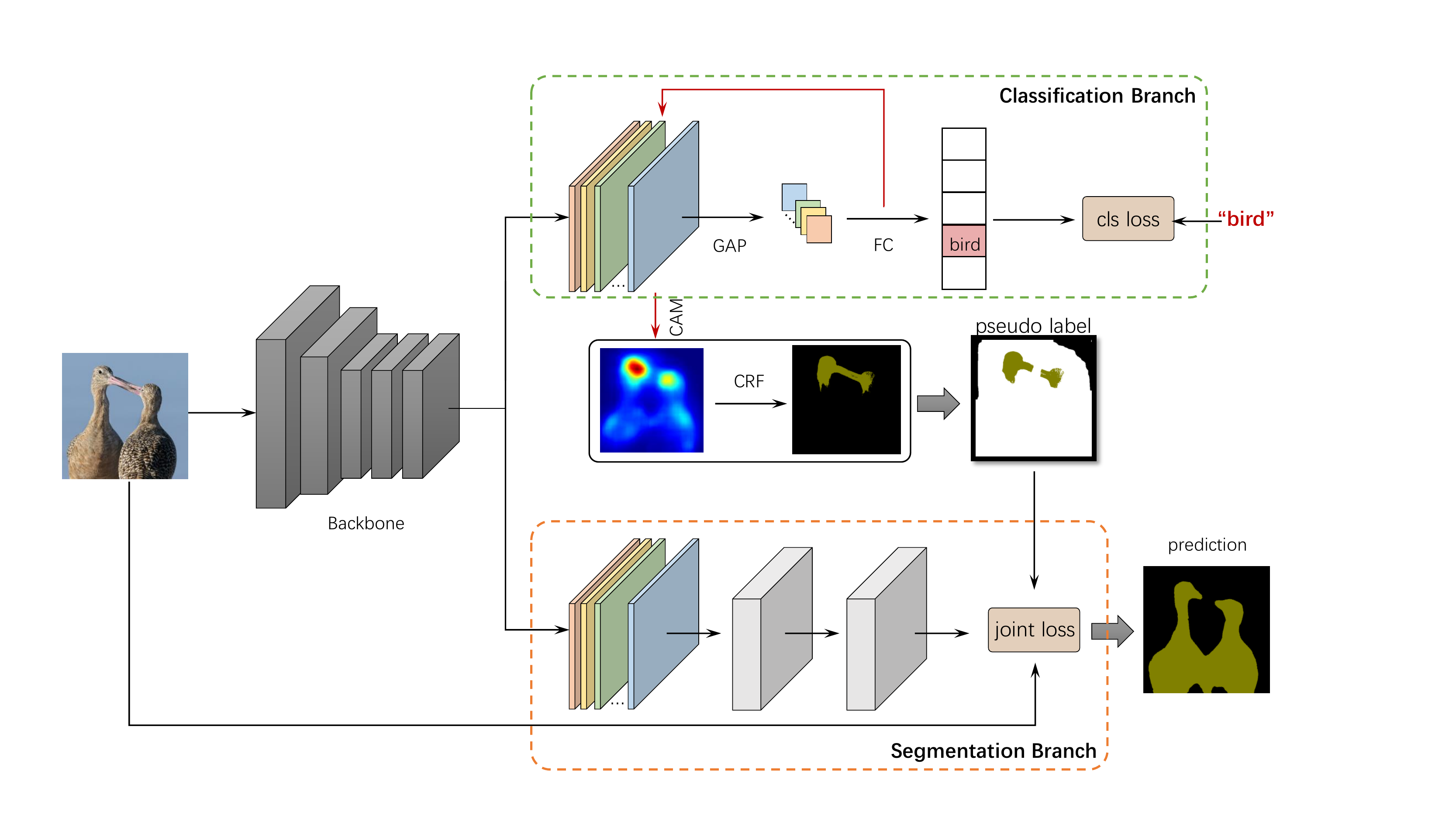}
	\caption{The framework of our proposed \emph{RRM} network. First of all, original regions are calculated through the classification branch, then the pseudo pixel-level masks are generated. Finally, the pseudo labels are applied as supervision to train the semantic segmentation branch. The whole RRM is jointly optimized end-to-end via a standard back-propagation algorithm during training.}
	\label{fig2}
\end{figure*}
\subsection{Overview}
Our proposed \emph{RRM} can be divided into two parallel branches including a classification branch and a semantic segmentation branch. Both branches share the same backbone network, and during training both of them update the whole network at the same time. The overall framework of our method is illustrated in Figure \ref{fig2}. The  algorithm flow is illustrated in Algorithm~\ref{alg:Framwork}.
\begin{itemize}
	\item The classification branch is used to generate reliable pixel-level annotations. Original CAMs will be processed to generate reliable yet tiny regions. The final remained reliable regions are regarded as labeled regions, while other regions are viewed as unlabeled. These labels are used as supervision information for the semantic segmentation branch for training. 
	\item The semantic segmentation branch is used to predict pixel-level labels. This branch deploys a new joint loss function combining the cross entropy loss with a newly designed dense energy loss. The cross entropy loss mainly considers labeled pixels, while the dense energy loss takes into account all pixels by making full use of RGB color and pixel positions. 
\end{itemize}

The overall loss function of our \emph{RRM} is: $\mathcal{L}=\mathcal{L}_{class} + \mathcal{L}_{joint_{-}seg} \label{Loss}$, where $\mathcal{L}_{class}$ represents a conventional classification softmax loss, while $\mathcal{L}_{joint_{-}seg}$ is a newly introduced joint loss for the segmentation branch.  

\subsection{Classification Branch: Generating Labels for Reliable Regions}\label{sec3.2}

High-quality pixel-level annotation has a direct impact on our final semantic segmentation performance as it is the only ground-truth in the training processing. Original CAMs can highlight the most discriminative regions of an object, but they still contain some non-object areas, which are the mislabeled pixels. Therefore, after getting the original CAM regions, post-processing such as dense CRF \cite{krahenbuhl2013parameter} is needed. We followed this basic idea and do further process for generating the reliable labels.

We compute the initial CAMs of the training dataset following~\cite{zhou2016learning}. In our network, global average pooling (GAP) is applied on the last convolution layer. The output of GAP is classified with a fully-connected layer. Finally, the fully-connected layer weights are used on the last convolution layer to obtain the heatmap for each class. Besides, inspired by the fact that dilated convolution can increase the respective field~\cite{wei2018revisiting}, we add dilated convolution into the last three layers. Details of our network settings are reported in our experiment Section.

Mathematically, given an image \emph{I}, the CAM of class $c$ is:
\begin{equation}
M_{ocam}^{c} = RS(\sum_{ch=0}^{D}\omega_{ch}^{c} \cdot F_{ch}), (c \in C_{fg}), \label{cam}
\end{equation}
where $C_{fg} = \left \{ c_1,c_2,...,c_N \right \}$ includes all foreground classes, $M_{ocam}^{c}$ is the CAM of class $c$ for image $I$,  $\omega^{c}$ denotes the weights of the fully-connected layer for class $c$, and \emph{F} is the feature maps from the last convolution layer of the backbone. $RS(\cdot)$ is an operation to resize the input to the shape of $I$. 

\begin{figure}[!htb]
	\centering
	\includegraphics[width=\columnwidth]{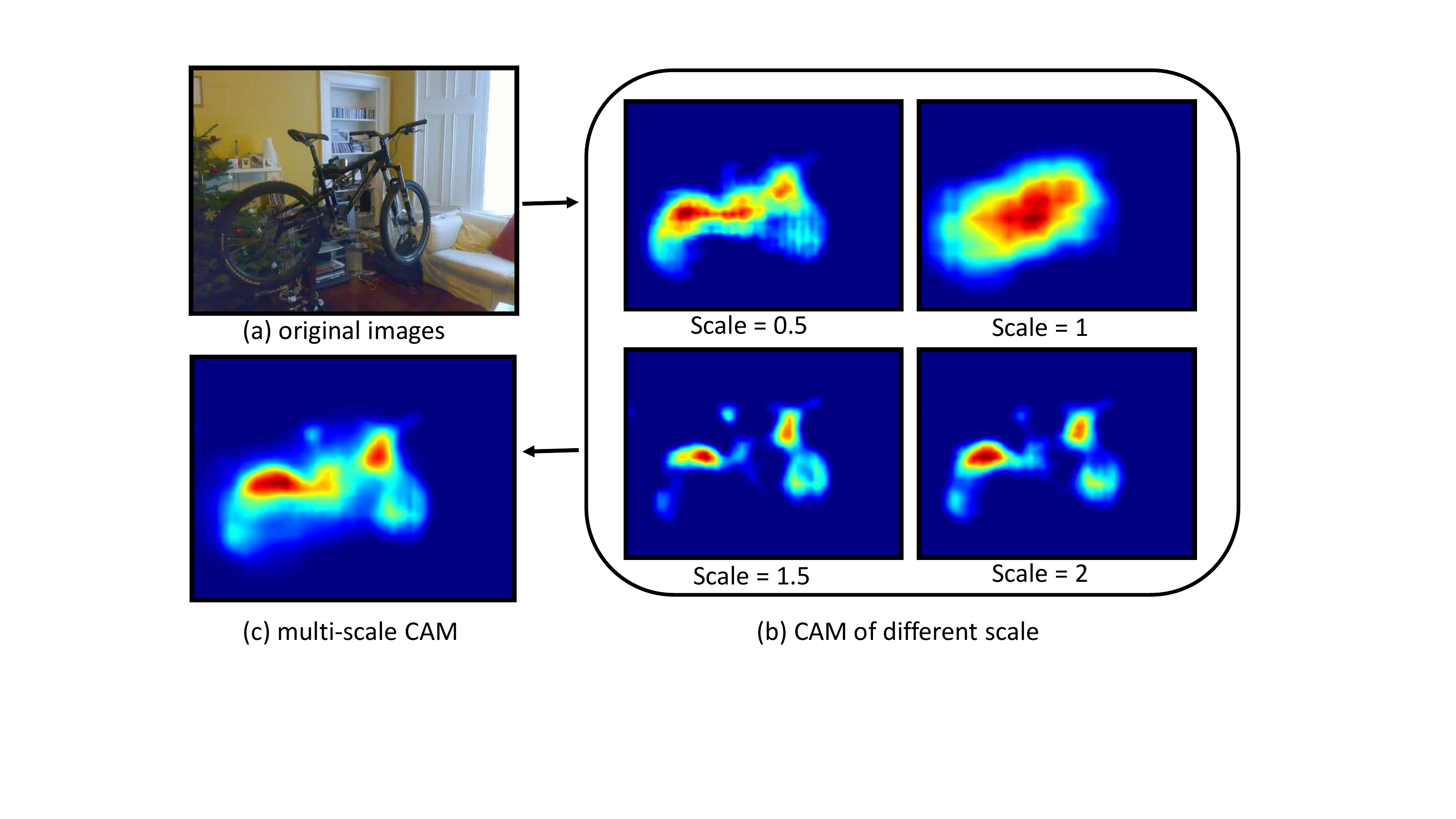}
	\caption{An example of computing multi-scale CAM.}
	\label{mcam}
\end{figure}

Using multi-scale of original images is beneficial for generating a stable CAM. Given $I$ and it is scaled by a factor $s_i$, $s_i \in \left \{ s_0,s_1,...,s_n \right \}$, the multi-scale CAM for $I$ is detonated as:
\begin{equation}
M_{cam}^{c} = \sum_{i=0}^{n} (M_{ocam}^{c}(s_i)/(n+1)),\label{eqcam2}
\end{equation}  
where $M_{ocam}^{c}(s_i)$ is the CAM of class $c$ for the scaled image $I$ with a factor $s_i$. Figure \ref{mcam} shows that compared to original CAM (scale=1), the multi-scale CAM provides more accurate object localization.

The CAM scores are normalized, so that we can get the classification probabilities for each pixel in \emph{I},
\begin{equation}
P_{fg}^{c} = M_{cam}^{c}/max(M_{cam}^{c}), (c \in C_{fg}),\label{eq2}
\end{equation}
where $max(M_{cam}^{c})$ is the maximum value in the CAM of class $c_j$. 

The background score is calculated using a similar way as in  \cite{ahn2018learning}:
\begin{equation}
P_{bg}(i) = (1 - \max_{c \in C_{fg}}(p_{fg}^{c}(i))^{\gamma}, \gamma > 1.\label{eq3}
\end{equation}
where $i$ is the pixel position index, $\gamma$ is the decay rate which helps to suppress background labels. The overall probability map, namely $P_{fg\_bg}$, is obtained by concatenating foreground and background probabilities $P_{fg}$ and $P_{bg}$.

After that, we use the dense CRF \cite{krahenbuhl2013parameter} as post-processing to remove some mislabeled pixels, and the CRF pixel label map is:
\begin{equation}
I_{crf} = CRF(I,[P_{fg}, P_{bg}]). \label{eq6}
\end{equation}

The selected reliable CAM label is:
\begin{equation}
\resizebox{0.87\columnwidth}{!}{$
I_{cam}(i)=\left\{\begin{matrix}
\argmax\limits_{c \in C}(P_{fg\_bg}^{c}(i)), & \text{if } \max\limits_{c \in C}(P_{fg\_bg}^{c}(i)) > \alpha\\ 
255, & else
\end{matrix}\right.$}, \label{eq7}
\end{equation}
where $C= \left \{ c_0,c_1,...,c_N \right \}$ includes all classes and the background ($c_0$). 255 means the class label is not decided yet.

The final pixel label input to the semantic segmentation branch is:
\begin{equation}
I_{final}(i) = \left\{\begin{matrix}
I_{cam}(i), & \text{if }  I_{cam}(i) =I_{crf}(i) \\ 
255, & else 
\end{matrix}\right.  \label{eqIfinal}
\end{equation}

In (\ref{eq7}), $\max\limits_{c \in C}(P_{fg_{-}bg}^{c}(i)) > \alpha$ selects the highly confident regions. In (\ref{eqIfinal}), $I_{crf}(i) = I_{cam}(i)$ considers the CRF constrains. Taking this strategy, highly reliable regions as well as their labels can be obtained. The regions which are detonated as 255 in (\ref{eqIfinal}) are regarded as unreliable regions.

\begin{figure}[!htb]
	\centering
	\includegraphics[width=\columnwidth]{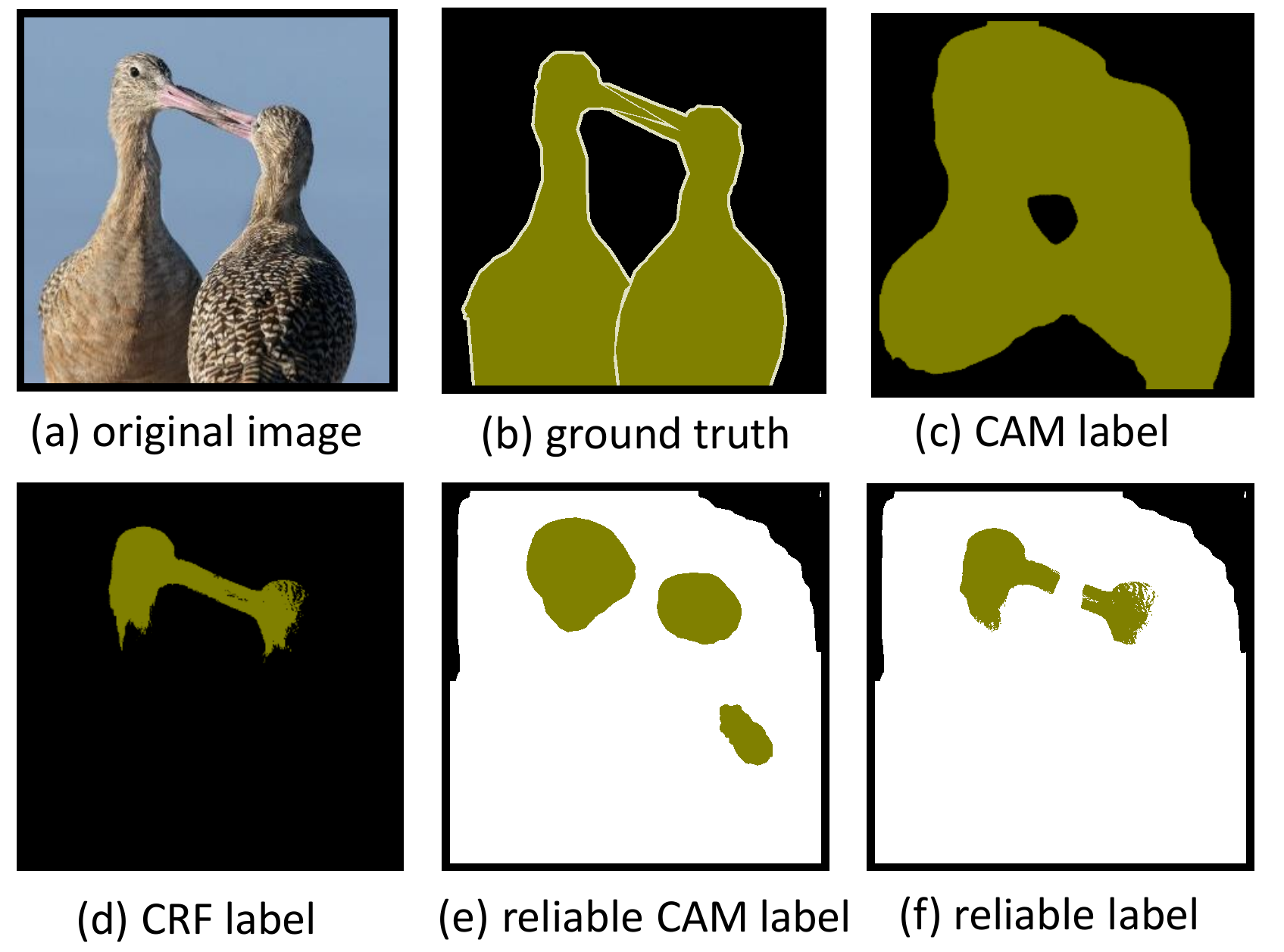}
	\caption{An example of generating reliable pixel labels. (c) is computed only considering the corresponding class label of $P_{fg\_bg}$. (d) is the result of (\ref{eq6}), (e) and (f) are generating through (\ref{eq7}) and (\ref{eqIfinal}), respectively. The white pixels in (e) and (f) are the unreliable regions.}
	\label{rcam}
\end{figure}

Figure~\ref{rcam} shows an example of our approach. It is observed that the original CAM labels ( Figure~\ref{rcam} (c)) contain most foreground labels but introduce a number of background pixels as foreground. The CRF label (Figure~\ref{rcam} (d)) can get accurate boundary but at the same time, many foreground pixels are regarded as background. In other words, the CAM label can provide reliable background pixels and CRF label can provide reliable foreground pixels. Combing the CAM label and CRF label map using our method, some wrong pixel-level labels are removed while the reliable regions are still remained, which is especially obvious at the object boundaries (see the difference between Figure~\ref{rcam} (e) and (f)).

\subsection{Semantic Segmentation Branch: Making Predictions }\label{sec3.3}

After getting the reliable pixel-level annotations, they are used as labels for our semantic segmentation branch. Different from the other methods which train their semantic segmentation network with the integral pseudo labels independently, our segmentation branch shares the same backbone network with the classification branch, needing only reliable yet tiny pixel-level labels. Our loss function consists of a cross entropy loss and a energy loss. Cross entropy loss focuses on utilizing the labeled data while the energy loss considers both labeled and unlabeled data. The joint loss is:
\begin{equation}
\mathcal{L}_{joint_{-}seg} = \mathcal{L}_{ce}+ \mathcal{L}_{energy} \label{eq8}.
\end{equation}
In (\ref{eq8}), $\mathcal{L}_{ce}$ and $\mathcal{L}_{energy}$ represent the cross entropy loss and the dense energy loss, respectively. The cross entropy loss is:
\begin{equation}
\mathcal{L}_{ce} = -\sum_{c \in C , i \in \Phi }B_{c}(i)log(P_{net}^{c}(i)),
\label{eq9}
\end{equation}
where $B_{c}(i)$ is a binary indicator, which equals to $1$ if the label of pixel $i$ is $c$ and otherwise $0$; $\Phi$ denotes the labeled regions, $\Phi  = \{i| I_{final}(i) \neq 255 \}$; $P_{net}^{c}(i)$ is the output probability of the trained network

So far, all labeled pixels has been used for training with cross entropy loss, but there are a large number of unlabeled  pixels. In order to make predictions for those unlabeled regions, we design a new shallow loss named dense energy loss considering both RGB colors and spatial positions.

We firstly define the energy formulation between pixel $i$ and $j$ based on \cite{joy2019efficient}:

\begin{equation}
E(i,j)=\sum_{\begin{subarray}{l}
	c_a,c_b\in C \\c_a \neq c_b
	\end{subarray}}
G(i,j)P_{net}^{c_a}(i)P_{net}^{c_b}(j).\label{Eij}
\end{equation}

In (\ref{Eij}), both $c_a$ and $c_b$ are the class labels, $P_{net}^{c_a}(i)$ and $P_{net}^{c_b}(j)$ are the softmax output of our segmentation branch at pixel $i$ and $j$, respectively. $G(i,j)$ is a Gaussian kernel bandwidth filter:
\begin{equation}
G(i,j)=\frac{1}{W}exp(-\frac{\left \| D_i-D_j \right \|^2}{2\sigma _{d}^{2}}-\frac{\left \| I_i-I_j \right \|^2}{2\sigma _{r}^{2}}),\label{Gij}
\end{equation}
where $\frac{1}{W}$ is the normalized weights,  $D$ is the pixel spatial position while $I$ is the RGB color. $\sigma_{d}$ and $\sigma_{r}$ are hyper parameters which control the scale of Gaussian kernels. (\ref{Eij}) can be simplified using Potts model \cite{tang2018regularized}:
\begin{equation}
\begin{aligned}
E(i,j)&=\sum_{\begin{subarray}{l}
	c_a,c_b\in C \\c_a \neq c_b
	\end{subarray}}
G(i,j)P_{net}^{c_a}(i)P_{net}^{c_b}(j) \\
&= G(i,j)\sum_{c\in C}P_{net}^{c}(i)(1-P_{net}^{c}(j)).\label{potts}
\end{aligned}
\end{equation}

Finally, our dense energy loss can be written as:
\begin{equation}
\mathcal{L}_{energy} = \sum_{i=0}^{N}\sum_{\begin{subarray}{l}
	j=0 \\j \neq i
	\end{subarray}}^{N}S(i)E(i,j). \label{scrf}
\end{equation}

In (\ref{scrf}), considering the fact that cross entropy loss is designed for supervised learning with label information 100\% accurate, but in this task, all pixel labels are not 100\% reliable, which means that using cross entropy loss might introduce some errors. Thus, our dense energy loss is applied to mitigate this problem. Based on this idea, we design a soft filter $S(i)$ for pixel $i$:

\begin{equation}
S(i) = \left\{\begin{matrix}
1- \max\limits_{c \in C}(P_{net}^{c}(i)), & i \in \Phi   \\  
1, & else
\end{matrix}\right.
\end{equation}

\begin{algorithm}[htb] 
	\caption{Algorithm flow  of our proposed approach.} 
	\label{alg:Framwork} 
		{\bf Input:} ~~Images $I$ with their image-level class labels $C_{fg}$;\\
		{\bf Output:}~~The trained end-to-end network, $Net$;
			
	\begin{algorithmic}[1] 		
		\WHILE{\emph{iteration} is \emph{true}}
		\STATE Use the classification Network branch to get the original CAMs;
    	\STATE Get the multi-scale CAMs with (\ref{eqcam2}) for each class;
		\STATE Use (\ref{eq2}) and (\ref{eq3}) to get foreground probability $P_{fg}$ and background probability $P_{bg}$;
		\STATE Get the overall CAM probability map $P_{fg\_bg}$ by combining  $P_{bg}$ and $P_{fg}$;
		\STATE Calculate reliable CAM label $I_{cam}$ and CRF label $I_{crf}$;
		\STATE Get the reliable regions and label $I_{final}$ from $I_{cam}$ and $I_{crf}$ using (\ref{eq7})(\ref{eqIfinal});
		\STATE Produce predictions and update the whole network using loss function $\mathcal{L} =  \mathcal{L}_{class} + \mathcal{L}_{ce} + \mathcal{L}_{energy}$;
		\ENDWHILE
		
	\end{algorithmic}
\end{algorithm}
\section{Experiments}
\subsection{Dataset and Implementation Details}\label{sec4}

\textbf{Dataset}. Our \emph{RRM} is trained and validated on PASCAL VOC 2012 \cite{everingham2010pascal} as well as its augmented data, including $10,582$ images for training, $1,449$ images for validating and $1,456$ images for testing. Mean intersection over union (mIoU) is considered as the evaluation criterion.

\noindent\textbf{Implementation Details}. The backbone network is a ResNet model with $38$ convolution layers~\cite{wu2019wider}. We remove all the fully connected layers of the original network and engage dilated convolution for the last three resnet blocks (a resnet block is a set of residual units with the same output size), the dilated rate is $2$ for the last third layer, and $4$ for the last $2$ layers. For the semantic segmentation branch, we add two dilation convolution layers of the same configuration after the backbone \cite{wu2019wider}, with kernel size $3$, dilated rate $12$, and padding size $12$. Cross entropy loss is computed for background and foreground individually. $\sigma_{d}$ and $\sigma_{r}$ in our dense energy loss are set as 15 and 100, respectively. 

The training learning rate is $0.001$ with weight decay being $5e$-$4$. The training images are resized with a ratio randomly sampled from $(0.7, 1.3)$, and they are randomly flipped. Finally, they are normalized and randomly cropped to size 321*321.

To generate reliable regions, the scale ratio in (\ref{eqcam2}) is set to $\{0.5, 1, 1.5, 2\}$, $\gamma$ in (\ref{eq6}) is set to $4$ for $P_{fg\_bg}$. The CRF parameters in (\ref{eq6}) follow the setting in \cite{ahn2018learning}. In (\ref{eq7}), an $\alpha$ value is chosen with $40\%$ pixels selected as labeled pixels for each class.  During validating and testing, dense CRF is applied as a post-processing method, and the parameters are set as the default values given in \cite{huang2018weakly}.
During training, both two branches update the backbone network. During testing, only the segmentation branch is used to produce the predictions.

\noindent\textbf{Reproducibility}: PyTorch \cite{paszke2017automatic} was used. All the experiments were performed on NVIDIA RTX 2080 Ti. Code now is available at: https://github.com/zbf1991/RRM.

\subsection{Analysis of Our Approach}

Our \emph{RRM} has two important aspects: using the reliable yet tiny pseudo masks for supervision and a new joint loss function for end-to-end training. Ablation studies are conducted to illustrate their individual and joint effectiveness, with results reported in Table \ref{camcrf1} and Table \ref{camdy}.

\begin{table}[!htb]
	\centering
	\resizebox{\columnwidth}{!}{
	\begin{tabular}{l|ccccccc}
		\hline
		Ratio & 0.1 & 0.2 & 0.3 & 0.4 & 0.6 & 0.8 & 1.0 \\ \hline
		CE loss & 0.486 & 0.487 & 0.484 & 0.485 & 0.483 & 0.495 & 0.557 \\ \cline{1-1} \hline
		Joint loss & 0.428 & 0.623 & 0.626 & 0.626 & 0.609 & 0.594 & 0.582 \\ \hline
	\end{tabular}
}
	\caption{Performance on PASCAL VOC 2012 \emph{val} set based on different mined region. Ratio means the proportion of reliable regions which is mined by our method to the whole pixels. "CE loss" means only cross entropy loss was used for our segmentation branch and "Joint loss" means our dense energy loss was combined with cross entropy loss was used for the segmentation branch. }\label{camcrf1}

\end{table}

We firstly validate the influence of different pseudo mask size. We do this by changing $\alpha$. Table \ref{camcrf1} reports the results. A smaller pseudo mask size means that more reliable regions are selected for the segmentation branch, while a larger size means that fewer reliable pixels are labeled. Table \ref{camcrf1} demonstrates that 20\%-60\% labeled pixels lead to the best performance. On one hand, too few labeled pixels cannot get satisfied performance since the segmentation network cannot get enough labels for learning. On the other hand, too many labeled pixels means more incorrect labels are used, which are noise for the training processing.

 \begin{table}[!htb]
 	\centering
 	\resizebox{0.6\columnwidth}{!}{
 	\begin{tabular}{l|cc}
 		\hline
 		& CE loss & Joint loss \\ \hline
 		CAM & 0.461 & 0.557 \\ \hline
 		\emph{Ours-RRM} & 0.485 & 0.626 \\ \hline
 	\end{tabular}
 }
 	\caption{Analysis of our method. CAM means class activate maps directly as pseudo masks. Ours-\emph{RRM} means that we used our method to produce pseudo masks. Both CAM and ours-\emph{RRM} use top 40\% pixels according to Table \ref{camcrf1}.}	\label{camdy}
 \end{table}
 
 \begin{table*}[!htb]
 	\centering
 	\resizebox{\textwidth}{!}{
 	\begin{tabular}{l|ccccccccccccccccccccc|c}
 		\hline
 		End-to-End Method       & bkg  & plane & bike & bird & boat & bottle & bus  & car  & cat  & chair & cow  & table & dog  & horse & mbk  & person & plant & sheep & sofa & train & tv   & mIOU \\ \hline
 		EM-Adapt \cite{papandreou1502weakly}    & 67.2 & 29.2  & 17.6 & 28.6 & 22.2 & 29.6   & 47.0 & 44.0 & 44.2 & 14.6  & 35.1 & 24.9  & 41.0 & 34.8  & 41.6 & 32.1   & 24.8  & 37.4  & 24.0 & 38.1  & 31.6 & 33.8 \\\hline
 		\emph{Ours-RRM (one-step)}  &  \textbf{87.9}     &    \textbf{75.9}   & \textbf{31.7}      &  \textbf{78.3}    &  \textbf{54.6}     &  \textbf{62.2}      & \textbf{80.5} & \textbf{73.7}    &\textbf{71.2}     & \textbf{30.5}  &  \textbf{67.4}      &  \textbf{40.9}     & \textbf{71.8}     &  \textbf{66.2}     &  \textbf{70.3}      & \textbf{72.6}      & \textbf{49.0}    & \textbf{70.7}      & \textbf{38.4}      &\textbf{62.7}     & \textbf{58.4}   & \textbf{62.6}      \\ \hline
 	\end{tabular}
 }
 	\caption{Performance on the PASCAL VOC 2012 \emph{val} set,  compared with other end-to-end weakly supervised approaches.}
 	\label{vocval}
 \end{table*}
 
 \begin{table*}[!htb]
 	\centering
 	\resizebox{\textwidth}{!}{
 	\begin{tabular}{l|ccccccccccccccccccccc|c}
 		\hline
 		End-to-End Method       & bkg  & plane & bike & bird & boat & bottle & bus  & car  & cat  & chair & cow  & table & dog  & horse & mbk  & person & plant & sheep & sofa & train & tv   & mIOU \\ \hline
 		EM-Adapt \cite{papandreou1502weakly}     & 76.3 & 37.1  & 21.9 & 41.6 & 26.1 & 38.5   & 50.8 & 44.9 & 48.9 & 16.7  & 40.8 & 29.4  & 47.1 & 45.8  & 54.8 & 28.2   & 30.0  & 44.0  & 29.2 & 34.3  & 46.0 & 39.6 \\ \hline
 		
 		\emph{Ours-RRM (one-step)} &  \textbf{87.8}    &    \textbf{77.5}  &    \textbf{30.8}  & \textbf{71.7}     & \textbf{36.0}     & \textbf{64.2}       &\textbf{75.3}    & \textbf{70.4}     &    \textbf{81.7}  & \textbf{29.3}      & \textbf{70.4}     & \textbf{52.0}     &\textbf{78.6}      & \textbf{73.8}      &\textbf{74.4}      &\textbf{72.1}        &\textbf{54.2}      &\textbf{75.2}       & \textbf{50.6}     &\textbf{42.0}       &\textbf{52.5}      &\textbf{62.9}      \\ \hline
 	\end{tabular}
 }
 	\caption{
 		Performance on the PASCAL VOC 2012 \emph{test} set, compared with other end-to-end weakly supervised approaches.}
 	\label{voctest}
 \end{table*} 
 
 \begin{table*}[!htbp]
 	\begin{center}
 		\resizebox{\textwidth}{!}{
 		\begin{tabular}{lllcccc}	
 			\hline
			 Method & Baseline & Sup. & Extra Data & End-to-end & val (mIoU) & test (mIoU) \\ \hline
 			Deeplab (ICLR'15) \cite{chen2014semantic} & VGG-16 &F & - & - & 67.6 & 70.3 \\
 			Deeplab-v2 \cite{chen2018deeplab} &ResNet-101 & F&- & - & 76.8 & 79.7 \\
 			\hline
 			WSSL (ICCV'15) \cite{papandreou1502weakly} & VGG-16& B & - & -& 60.6 & 62.2 \\
 			BoxSup (ICCV'15) \cite{dai2015boxsup} & VGG-16 &B &- &- & 62.0 & 64.6 \\
 			\hline
 			ScribbleSup (CVPR'16) \cite{lin2016scribblesup} & VGG-16 &S & - & - & 63.1 &-  \\
 			Kernel Cut (ECCV'18) \cite{tang2018regularized} & ResNet-101 &S &- & - & 75.0 &-  \\ 
 			\hline
 			CrawlSeg (CVPR'17) \cite{hong2017weakly} & VGG-16 & L& YouTube Videos & $\times$ & 58.1 & 58.7 \\
 			DSRG (CVPR'18) \cite{huang2018weakly} & VGG-16 & L& MSRA-B &$\times$ & 59.0 & 60.4 \\
 			DSRG (CVPR'18) \cite{huang2018weakly} & ResNet-101 &L & MSRA-B &$\times$ & 61.4 & 63.2 \\
 			FickleNet (CVPR'19) \cite{lee2019ficklenet}  & ResNet-101 &L & MSRA-B &$\times$ & 64.9 & 65.3 \\ 
 			\hline
 			EM-Adapt (ICCV'15)\cite{papandreou1502weakly} & VGG-16 &L & - &\checkmark & 38.2 & 39.6 \\
 			SEC (ECCV'16) \cite{kolesnikov2016seed} & VGG-16 &L &- &$\times$ & 50.7 & 51.7 \\
 			AugFeed (ECCV'16) \cite{qi2016augmented} & VGG-16 &L & - & $\times$& 54.3 & 55.5 \\
			AdvErasing (CVPR'17) \cite{wei2017object} & VGG-16 &L & - & $\times$& 55.0 & 55.7 \\
 			AffinityNet (CVPR'18) \cite{ahn2018learning} & VGG-16 &L &- &$\times$ & 58.4 & 60.5 \\  
 			AffinityNet (CVPR'18) \cite{ahn2018learning} & ResNet-38 &L &- &$\times$ & 61.7 & \textbf{63.7} \\  
 			\hline
 			\emph{Ours-RRM-VGG (two-step)}  & VGG16 &L & - & $\times$ & \textbf{60.7}  & \textbf{61.0}\\ 
 			\emph{Ours-RRM-ResNet (two-step)}  & ResNet-101 &L& - & $\times$ & \textbf{66.3}   & \textbf{66.5}\\ 
 			\emph{Ours-RRM (one-step)}  & ResNet-38 &L& - & \checkmark & \textbf{62.6} & 62.9  \\ 
 			\hline
 		\end{tabular}
 	}
 		\caption{
 			Comparison with the state-of-the-art approaches on PASCAL VOC 2012 \emph{val} and \emph{test} dataset. 
 			Sup.-supervision information, GT-ground truth, F-full supervision,  L-image-level class label, B-bounding box label, S-scribble label.
 		}\label{tab-state-of-art}
 	\end{center}
 \end{table*}

 Table \ref{camdy} shows the effectiveness of our introduced two main parts: reliable region mining and the joint loss. The results obtained using original CAM regions and the mined reliable regions with \emph{RRM} are compared. It is observed that the pseudo label generated by \emph{RRM} outperforms CAM labels. If we remove the joint loss from our segmentation branch, it also shows that the reliable pseudo labels generated by \emph{RRM} improves the segmentation performance. 

In addition, the comparison between \emph{Ours-RRM} with CE loss and \emph{Ours-RRM} with Joint loss in Table \ref{camdy} illustrates the effectiveness of the introduced joint loss. Without the joint loss, the mIoU obtained with \emph{RRM} with CE loss gets lower. This is because the mined reliable regions with \emph{RRM} cannot provide sufficient labels for segmentation model training when only considering cross entropy loss. After adopting the joint loss, segmentation performance improves with a big margin from 48.5 to 62.6, which is a 14.1 increase. Similar comparison result is obtained between \emph{CAM} with CE loss and \emph{CAM} with Joint loss.

\subsection{Comparisons with Previous Approaches}

In Table \ref{vocval} and Table \ref{voctest}, we make detailed comparisons with other end-to-end network for image-level-only supervised semantic segmentation. Although there are various different networks for this task, only EM-Adapt \cite{papandreou1502weakly} adopts an end-to-end structure, and it can be seen that \emph{Ours-RRM (one-step)} outperforms it with a big margin. First of all, compared with EM-Adapt \cite{papandreou1502weakly}, which uses an expectation–maximization (EM) algorithm to update the network parameters, our method adopts a more direct and explicit learning procedure to update the whole network, using our designed joint loss function. Secondly, EM-Adapt \cite{papandreou1502weakly} can only give a rough segmentation result as only the image-level information is considered, while \emph{Ours-RRM (one-step)} designs a pilot mechanism to provide reliable pixel-level labels, which leads to more accurate segmentation predictions.

In order to show the effectiveness and scalability of our idea, we also extend our method to a two-step framework. The difference is that for our one-step method 
(\emph{Ours-RRM (one-step)}), we produce the predictions through our segmentation branch directly. Whereas for our two-step method, we firstly used our \emph{Ours-RRM (one-step)} network to produce the pseudo masks for the training dataset.
Following that, we train and evaluate Deeplab \cite{chen2014semantic} with those generated pixel labels, which is named as \emph{Our-RRM-VGG (two-step)}. Using the same setting, we also evaluate the performance when Deeplab-v2 \cite{chen2018deeplab} with ResNet-101 backbone was used, called \emph{Our-RRM-ResNet (two-step)}.
The final results can be found in Table \ref{tab-state-of-art}. It is observed that among existing methods solely using image-level label without extra data, AffinityNet \cite{ahn2018learning} is the most performing one. However, both \emph{Ours-RRM-VGG (two-step)} and \emph{Ours-RRM-ResNet (two-step)} perform much better than it when the same backbone was used. One more thing should be noticed is that AffinityNet \cite{ahn2018learning} used ResNet-38 \cite{wu2019wider} as baseline, which is more powerful than ResNet-101 \cite{lee2019ficklenet}, and even in this case \emph{ours-RRM-ResNet (two-step)} still outperforms it with a big margin. 
Note that AffinityNet \cite{ahn2018learning} applies three different DNNs with many bells and whistles, while we get equivalent results with only one end-to-end network (\emph{Ours-RRM (one-step)}).  

To the best of our knowledge, the previous state-of-the-art, FickleNet \cite{lee2019ficklenet},  achieves the mIoU score of 64.9 and 65.3 on PASCAL VOC \emph{val} and \emph{test} set, but it uses class agnostic saliency map \cite{liu2010learning} as extra supplement information and uses two individual networks separately. \emph{Ours-RRM-ResNet (two-step)} gives a better performance with mIoU scores of 66.3 and 66.5 on PASCAL VOC \emph{val} and \emph{test} set, which represents 1.4 and 1.2 improvement. Note that we do not use extra data or information in our case. Therefore, \emph{Ours-RRM-ResNet (two-step)} is the new state-of-the-art for two-step image-level label weakly supervised semantic segmentation. 

\begin{figure}[hb]
 	\centering
 	\includegraphics[width=\columnwidth]{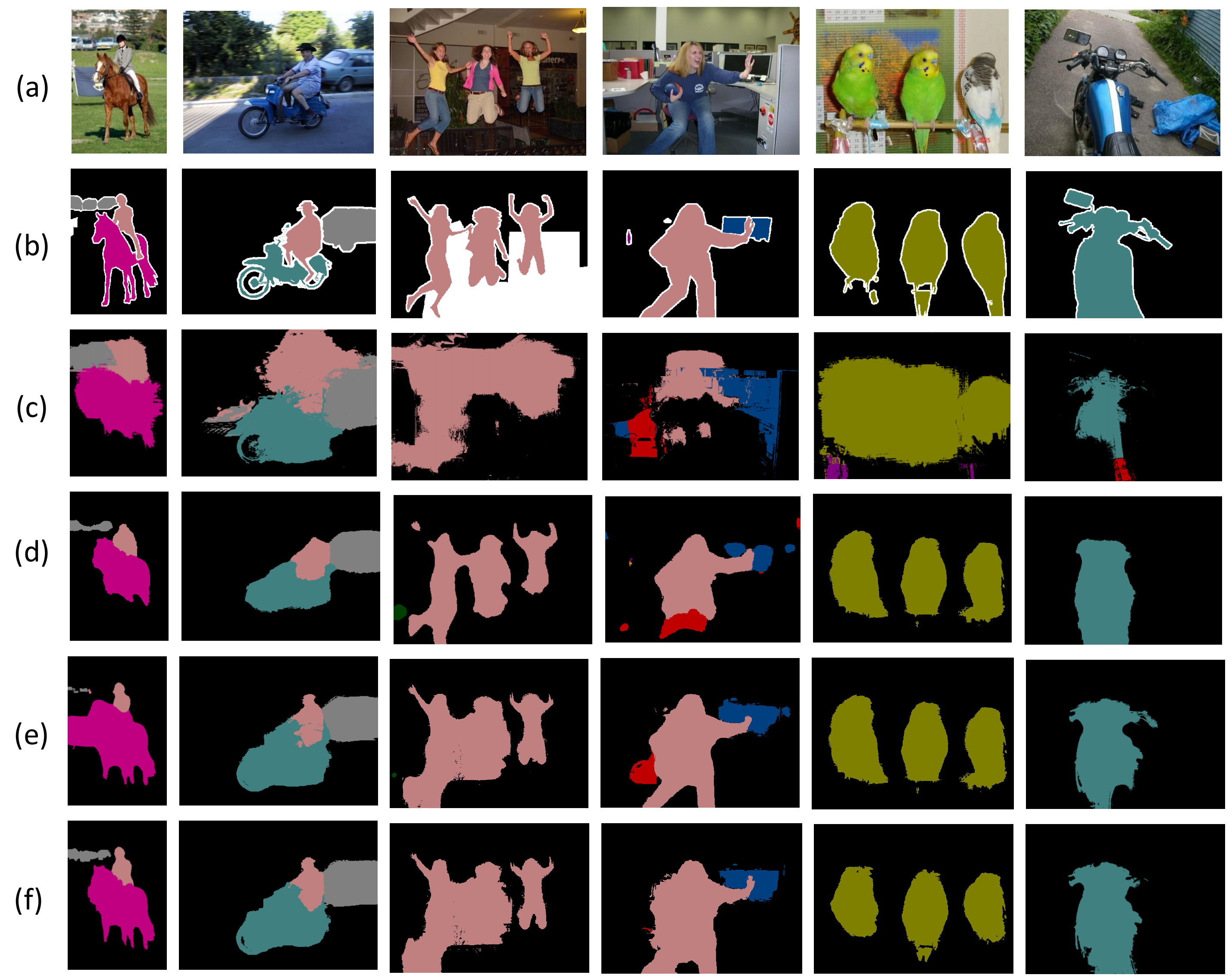}
 	\caption{Qualitative segmentation results on PASCAL VOC 2012 \emph{val} set. (a) Original images. (b) Ground-truth. (c) EM-Adapt results. (d) \emph{Ours-RRM (one-step)} results. (e) \emph{Ours-RRM-VGG (two-step)} results. (f) \emph{Ours-RRM-ResNet (two-step)} results.}
 	\label{vis}
 \end{figure}

In Figure \ref{vis}, we report some subjective semantic segmentation results of ours methods, which are compared with EM-Adapt \cite{papandreou1502weakly}, the state-of-the-art end-to-end network.  \emph{Ours-RRM (one-step)} obtains much better segmentation results on both large and small objects, with much accurate boundaries. We also show some results of our two-step approaches, and it can be seen that among our three methods, \emph{ours-RRM-ResNet (two-step)} obtains the best performance duo to the powerful network architecture.
\section{Conclusion}

In this paper, we proposed the \emph{RRM}, an end-to-end network for image-level weakly supervised semantic segmentation. We revisited drawbacks of the state-of-the-arts, which adopt the two-step approach. We proposed a one-step approach through mining reliable yet tiny regions and used them as ground-truth labels directly for segmentation model training. 
With limited pixels as supervision, we designed a new loss named dense energy loss, which takes shallow features (RGB colors and spatial information) and cooperates with the pixel-wise cross-entropy loss to optimize the training process. Based on our one-step \emph{RRM}, we extend a two-step method. Both our one-step and two-step approaches achieve state-of-the-art performance. More importantly, our \emph{RRM} offers a different perspective from the traditional two-step solutions. We believe that  the proposed one-step approach could further boost research in this direction.

\section{Acknowledgment}
The work was supported by National Natural Science Foundation of China under 61972323 and 61876155, and Key Program Special Fund in XJTLU under KSF-T-02, KSF-P-02, KSF-A-01, KSF-E-26.

\bibliographystyle{./aaai}
\bibliography{./ref}
\end{document}